\documentclass[11pt,a4paper]{article}
\usepackage{times}
\usepackage{latexsym}
\usepackage{graphicx}
\usepackage{amssymb}
\usepackage{amsmath}
\DeclareMathOperator*{\Norm}{Norm}
\DeclareMathOperator*{\Count}{Count} \DeclareMathOperator*{\Nw}{Nw}
\DeclareMathOperator*{\Logit}{Logit}
\DeclareMathOperator*{\Expit}{Expit}

\usepackage{url}
\usepackage{xcolor}

\title{Relationship Between Online Harmful Behaviors and Social Network Message Writing Style}

\author{Talia Sanchez Viera \\
Department of Computer Science and Software Engineering \\
 Universit\'e Laval \\
 Qu\'ebec, Canada \\
 {\tt talia.sanchez-viera.1@ulaval.ca} \\
 Richard Khoury\thanks{Corresoponding author} \\
 Department of Computer Science and Software Engineering \\
 Universit\'e Laval \\
 Qu\'ebec, Canada \\
 {\tt richard.khoury@ift.ulaval.ca} \\
 }

\begin{document}
\maketitle
\begin{abstract}
In this paper, we explore the relationship between an individual's writing style and the risk that they will engage in online harmful behaviors (such as cyberbullying). In particular, we consider whether measurable differences in writing style relate to different personality types, as modeled by the Big-Five personality traits and the Dark Triad traits, and can differentiate between users who do or do not engage in harmful behaviors. We study messages from nearly 2,500 users from two online communities (Twitter and Reddit) and find that we can measure significant personality differences between regular and harmful users from the writing style of as few as 100 tweets or 40 Reddit posts, aggregate these values to distinguish between healthy and harmful communities, and also use style attributes to predict which users will engage in harmful behaviors.
\end{abstract}

\section{Introduction}\label{intro}
Social networks and online communities are an important part of life for over a billion people worldwide, allowing unprecedented levels of social interactions, access to news, political involvement, and more. The personality of users in these virtual communities reflect their real-world personalities, meaning most are decent and respectful people, but some are harmful individuals with malicious intentions \cite{hardaker2010trolling}, and their actions online have negative impacts on other users and on the community overall \cite{mohan2017impact}. Most communities try to limit the negative impact of these individuals through moderation, by detecting harmful messages that have been posted and deleting them. However, this solution is inherently insufficient: since it makes no prediction on which users will post harmful messages, moderators must necessarily monitor the entire social network at all times or rely on other users to report harmful comments.

In this paper, we follow on recent studies that show it is possible to correlate a user's personality (as determined by a self-reporting questionnaire) to their writing style. We demonstrate that it is possible to use this correlation to predict a user's personality based on their writing style, and then use this prediction to pinpoint which users are likely to post harmful comments. We can further determine quantitatively which personality traits (from the Big-Five and Dark Triad) and which features of writing style (from the LIWC) are indicative of safe or risky users, and aggregate results over an entire group of users to show the health of a community. In practice, we believe this predictive ability would be of great use to community moderators, allowing them to focus their efforts on potentially harmful individuals and groups.

The rest of this paper is structured as follows. In the next section, we give a brief background of the main concepts in this paper and a review of the relevant literature. In section \ref{methodology}, we describe the methodology we use to predict and quantify users' personality traits. We apply this methodology in two sets of experiments, which we will present and analyze in section \ref{experimentalresults}. Section \ref{toxicityregression} presents the main contribution of our paper, the use of regression algorithms to predict the harm level of users based on their writing-style-inferred personality traits. In section \ref{danger}, we discuss the ethical considerations surrounding our work. Finally, section \ref{conclusion} will present our conclusions and suggest directions for future works.

\section{Related Work}\label{relwork}
Modern psychology commonly models human personality on five scales, which are called the Big-Five personality traits or the OCEAN model \cite{sumner2012predicting,de2000big}.
These five traits are:

\begin{itemize}
\item Openness (Op) to new experiences.
\item Conscientiousness (Co), indicating whether the person tends to be more organized and structured (or stubborn at the extreme), or spontaneous and care-free (or sloppy at the extreme).
\item Extraversion (Ex), or its lack (aka introversion).
\item Agreeableness (Ag), which is the importance that the person gives to getting along with others. A highly agreeable person will be very friendly, while a person with low agreeableness will be more detached and solitary.
\item Neuroticism (Ne), a measure of emotional stability in the face of anxiety and depression.
\end{itemize}

To these five positive traits, recent studies have added three negative personality traits to describe antisocial and amoral behavior \cite{paulhus2002dark,paulhus2011dark,medjedovic2015dark}. This ``Dark Triad'' of personality traits is composed of:

\begin{itemize}
\item Narcissism (Na), a need for self-aggrandizement and admiration along with a lack of empathy for others. 
\item Machiavellianism (Ma), a remorseless ability to manipulate others and exploit their weaknesses achieve one's own goals.
\item Psychopathy (Ps), deriving enjoyment from attacking, abusing, and intimidating those weaker than themselves.
\item Sadism is a fourth trait that is sometimes added to the previous three, to form a Dark Tetrad instead of a Dark Triad. It represents the personality trait that finds pleasure in the suffering of others.
\end{itemize}

The authors of \cite{sumner2012predicting} explored the relationship between language use on Twitter and personality traits. They did so by selecting 3,000 volunteers who completed a standard personality questionnaire to accurately measure their ranking in each of the eight traits of the model. Next, the authors obtained each volunteer's tweet history and measured the linguistic attributes defined in the Linguistic Inquiry and Word Count (LIWC) software, which range from pronoun and punctuation usage to semantic categories. Their results highlighted some interesting positive and negative correlations between personality traits and linguistic attributes. For example, they observed that the usage of first-person-plural pronouns is positively correlated with extroversion and agreeableness and negatively correlated with Machiavellianism and psychopathy, and that the usage of perceptual process words like ``heard'' or ``feeling'' is positively correlated with agreeableness but negatively correlated with neuroticism and psychopathy. Table \ref{tbl:corr-example} shows these sample correlation values.

\begin{table*}[t!]
\begin{center}
\small
\begin{tabular}{ccccccccc}
\hline
 Category & Na & Ma & Ps & Op & Co & Ex & Ag & Ne\\
\hline
1st-person plural & 0.036 & -0.070 & -0.071 & -0.006 & 0.052 & 0.063 & 0.068 & 0.044\\
Perceptual process & -0.052 & -0.040 & -0.092 & -0.015 & -0.020 & -0.031 & 0.070 & -0.089\\
\hline
\end{tabular}
\end{center}
\caption{Example of Spearman's correlations between linguistic variables and personality scores from \cite{sumner2012predicting}.}
\label{tbl:corr-example}
\end{table*}

Much like \cite{sumner2012predicting}, the authors of \cite{golbeck2011predicting} conducted a study focusing on the prediction of the Big-Five traits in Twitter users using 14 linguistic variables. The correlations they observed were for the most part consistent with those of \cite{sumner2012predicting}, although differences did exist, especially for the prediction of the neuroticism trait. 

The authors of \cite{preotiuc2016studying} conducted a study of the correlation between observed behaviors on Twitter and self-reported Dark Triad scores. They also used the linguistic categories of the LIWC and added some new features such as the user's profile image. Their study also highlighted a correlation between linguistic attributes and Dark Triad personality traits, confirming the results of \cite{sumner2012predicting}. The authors of \cite{schwartz2013personality} performed a similar study on Facebook, using both LIWC categories and an open vocabulary, and found correlations with self-reported scores of the Big-Five personalities, as well as with physical attributes such as gender and age. 

An alternative to the Big-Five model is the Schwartz's 10-value model, which was used in \cite{boyd2015values}. The authors of that study explored the relationship between language in Facebook status updates and these personality values. Using the MEM mapping of words to 30 different themes and the self-reported Schwartz's value survey, they were able to correlate language use to the Schwartz's model. They found that a correlation does exist between the two, but it is weaker than the authors expected, a fact which they say is indicative of the mismatch between the theoretical categories of values of the Schwartz's model and the real concerns people express on Facebook. They further indicate that a simple word-count of value keywords gives a better prediction of a person's values than the self-reported questionnaire. 

The challenge of personality prediction from social network data has also gathered interest from the data mining and natural language processing communities. We refer the reader to \cite{mehta2020recent} for a thorough review of that research area from the 1990s to the late 2010s.
In some newer developments, the authors of \cite{arnoux201725} conducted a study predicting self-reported Big-Five traits from Twitter trigrams, word embeddings and LIWC categories. The authors of \cite{christian2021text} used a set of deep-learning architectures and pre-trained language models to successfully predict personality traits from social network message texts, using the evaluation of a psychology expert as a baseline. 
Personality prediction from longer text documents has also risen in popularity recently, most notably using the Essays dataset \cite{pennebaker1999linguistic}, a dataset of stream-of-consciousness essays paired with the authors' Big-Five personality traits. The authors of \cite{kazameini2020personality} trained a resource-light deep architecture to predict personality traits from the essays, while \cite{jiang2020automatic} showed that using pre-trained contextual embeddings and attention networks improves results, and \cite{tinwala2021big} trained a convolution neural network for that task. In all cases, these authors concluded that the use of deep learning improves the personality prediction from textual features. The same observation was made by \cite{mehta2020recent}.

Finally, we can note that other studies have expanded the scope to prediction using a user's entire digital footprint \cite{azucar2018predicting,souri2018personality}. This gives the system a much richer source of information to base its predictions on, including elements such as \textit{likes}, uploaded materials, number of friends, reactions from friends, or counters of specific interests (e.g. number of sports teams liked). For example, \cite{utami2021personality} scrapped 8 activity features from the Facebook profiles of 170 volunteers who also completed a personality questionnaire, and trained an SVM to classify the profiles into the Big-Five personality traits. Other authors have incorporated Twitter metadata \cite{jeremy2019identifying}, Instagram pictures \cite{ferwerda2018predicting,ferwerda2018you}, video game profiles \cite{potard2020video}, online role-playing personae \cite{delhove2020relationship}, and much more. However, our study is designed to use only message text and what can be quantified about writing style from that text. Studies such as those cited above, which use very different and much richer input, are outside of our scope.

\subsection{Research Contributions}
This paper makes two key contributions.

The previous review highlights a common methodology to research projects in this area: they all rely on labelled data, most commonly self-reported questionnaires filled in honestly by participants \cite{sumner2012predicting,preotiuc2016studying, boyd2015values,arnoux201725} or occasionally expert opinions \cite{christian2021text} or labelled datasets \cite{kazameini2020personality,jiang2020automatic}, in order to learn  in a supervised manner how writing style is correlated to or predictive of personality traits. We propose instead to build on existing results in the literature to create an algorithm to infer personality traits from writing style in an unsupervised way. 
This means that our proposed methodology does not require users to submit honestly-completed self-reported personality questionnaires, an important departure from the existing literature.


Furthermore, we will give a particular focus to harmful users, which are the ones moderators would need to monitor closely to protect their online communities. These users are easy to identify after the fact, by using standard harmful-message detection tools. In this study, we will explore how the characteristics of their written messages correlate to different personality traits, and which characteristics are predictive of harmful online behaviors. Online harm is a major issue for online communities and has serious real-world consequences on individuals targeted, including leading to suicide \cite{adlreport}.
Yet, to the best of our knowledge, the only previous work studying the link between personality traits and online harm is that of \cite{buckels2014trolls}, and it was limited specifically on trolling. By administering personality questionnaires to hundreds of volunteers and correlating the results to their self-declared favorite online activities, the authors found a strong correlation between sadism and trolling, and even that sadism was a predictor of trolling behavior. On the other hand, they also found that narcissism, another Dark trait, was negatively correlated with trolling and positively correlated with enjoying participating in online debates, a positive behavior. 
By contrast, our study considers a much wider scope of harmful behaviors, and tackles the challenge of predicting harmful users in an unsupervised manner replicating the conditions found in real online communities. 

\section{Inferring Personality Traits}\label{methodology}
The first objective of this work is to explore an unsupervised technique to infer personality traits from writing style. We base our approach on existing results in the literature; more specifically, we chose to use the work of \cite{sumner2012predicting} as a starting point, as it was the most complete and readily available. In that paper, the authors use self-declared personality tests and compute language statistics from tweets in order to discover correlations between writing style and personality traits. Our approach does the opposite: we compute language statistics from online messages and apply correlations to infer personality traits .

More specifically, to model a user's personality from their messages, the first step is to gather the set of their most recent messages sent to the online community. We then sum the occurrences of each linguistic feature described in \cite{sumner2012predicting} over these messages. To do this, we use the Linguistic Inquiry and Word Count (LIWC) software version 2007, which was the same version used by \cite{sumner2012predicting}.
 
Once we have the number of occurrences of each feature for each user, we need to normalize these count values to make it possible to compare multiple users. The reason is that a simple counter will be biased towards users who write more. For example, a user who wrote more and longer messages will naturally use more first-person plural pronouns than one who wrote less, and since that feature is positively correlated to agreeableness and negatively correlated to psychopathy that user will appear more agreeable and less psychopathic than the one who wrote less. We considered normalizing by the number of messages, but again we found this would introduce a bias in our values, this time in favor of users who write many short messages rather than one long one. In addition, it would create a bias for platforms such as Twitter, which enforce a message length limit and thus force users to split a long message into multiple short messages, over other platforms that do not have that artificial restriction. Consequently, we normalize by using the total length of all recent messages, which we compute as the total number of written words.

Next, a second normalization step is necessary. The reason is that certain linguistic features, such as punctuation and personal pronouns, are used much more frequently than others, such as personal concern (work-related or death-related) words. As a result, the more popular features eclipse the less-popular ones in value, and they end up exclusively defining the personality of the user while the less-popular features have no impact. We deal with this issue by implementing a simple min-max normalization of the values of each feature across all users. This normalization performs a linear transformation on the occurrence values of each feature so that the minimum and maximum value of each feature is 0.0 and 1.0 respectively and that the ordering and ratios between users are maintained. For completeness, the equation is: 

\begin{equation}
\Norm(x) = \frac{x - \min_x}{\max_x - \min_x},
\end{equation}

\noindent where $x$ is the original occurrence value of a feature for a user, and $\min_{x}$ and $\max_{x}$ are the minimum and maximum observed occurrence value of that feature, respectively. We could have used a z-score normalization instead, but opted for min-max simply because it is more commonly used in practice.

We decided to use the correlation values of \cite{sumner2012predicting} as weights that quantify the relative importance of each language feature for each personality trait. We can thus compute the value of a personality trait for a given user as a weighted sum of their observed features:

\begin{equation}\label{eq2.2}
P_{i,u} = \sum_{f=0}^{F} w_{i,f} \times \Norm\left( \frac{\Count_{u,f}}{\Nw_u} \right),
\end{equation}

\noindent where $P_{i,u}$ is personality trait $i$ (one of the eight defined in Section \ref{relwork}) as computed for user $u$, $w_{i,f}$ the weight of linguistic feature $f$ for computing personality trait $i$ (the correlation between feature $f$ and trait $i$), and $\frac{\Count_{u,f}}{\Nw_u}$ is the number of times feature $f$ was observed in the messages sent by user $u$ normalized by the total number of words written by that user.

\section{Personality Inference Results}\label{experimentalresults}

\subsection{Twitter Experiment}\label{experimentTwitter}
\begin{figure}[htbp]
\centering
\includegraphics[scale=0.5]{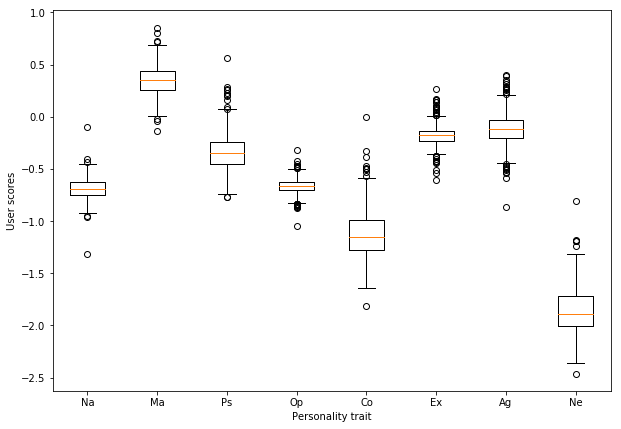}
\caption{Boxplot representation of personality traits in Twitter dataset.}
\label{fig:box_twitter8} 
\end{figure}

Since the work of \cite{sumner2012predicting}, which our work is based on, studied Twitter messages, we decided to conduct a first experiment on that social network. We created a dataset by randomly selecting some 850 users who had public profiles \footnote{While we used exclusively public data, we opt not to share our dataset to preserve the anonymity of the individuals whose profiles were randomly sampled.}, between 350 and 400 tweets posted, and used exclusively English. We computed the eight personality trait scores of each user with the method described in Section \ref{methodology}. The spread of values for each personality trait is shown in the boxplot of Figure \ref{fig:box_twitter8}. Overall, the short span of the interquartile range (IQR) in our results indicate that most users have personality values within a small range of each other. Some of the results of specific traits were expected, such as high extraversion and high agreeableness which indicate that the users enjoy talking and interacting with others. Others, especially the high Machiavellianism and psychopathy scores, were unexpected.

For each personality trait in Figure \ref{fig:box_twitter8}, users inside the boundaries of the boxplot are ``normal'' users\footnote{These users are ``normal'' in the sense that they show a level of a trait that is consistent with the majority of the user-base, not in a healthcare or psychological sense, which we cannot evaluate in this work.}, while outlier users above and below the boxplot show respectively more or less than a normal level of that trait. We thus decided to compare the messages of users inside the boxplot to those of outlier users above and below the plot for each personality trait.

To evaluate and analyse these results, we need to compare them to an objective baseline. Many of the studies presented in Section \ref{relwork} use the results of self-reported psyschological tests for that purpose, but that is not an option for our study. Instead, we need to label the tweets for harmful content. However, manual labeling by crowdsourcing services risks introducing biases in the data \cite{barbosa2019rehumanized}, while harm detection systems in the literature are limited to the specific types of harm and specific communities they are trained on \cite{larochelle2020generalisation}. Consequently, we opt to use a commercial general-purpose harm labeling system, namely the Community Sift system by Two Hat Security. 
For this study, we will focus on five types of harm identified by the system: general risk, bullying, fighting, vulgar, and sexting. For each type, the software rates comments on an eight-level risk scale, where levels 0 to 3 represent low-risk comments from ``super-safe'' to ``questionable'', levels 5 to 7 represent high-risk comments from ``mild'' to ``severe'', and level 4 represents ``unknown'' comments (usually gibberish) that cannot be rated. For our research, we ignore level 4 as uninformative. 

We can compare the ratio of tweets sent at each level of harm by users in each personality group. Table \ref{tbl:Perc-toxicMess-Twitter} shows the results obtained. Looking at our results for the three Dark Triad traits, it can be seen that outlier users who have above-normal levels of these traits have a considerably lower ratio of safe messages (levels 0 to 3) and a higher ratio of unsafe messages (levels 5 to 7) compared to users inside the boxplot. On the other hand, outlier users with below-normal levels of all three Dark Triad traits show the opposite behavior, a higher ratio of safe messages and a lower ratio of unsafe messages. While the results for psychopathy and Machiavelism are coherent with others found in the literature, the results for narcissism are at odds with the literature, most notably \cite{preotiuc2016studying} who observed that messages from high narcissism users were correlated with positive emotions, and \cite{buckels2014trolls} who report that narcissism and trolling behaviors are not correlated. This difference likely comes from the wider scope of our benchmark: \cite{preotiuc2016studying} only considered LIWC categories and \cite{buckels2014trolls} focused on a single harmful behavior, while our benchmark software detects a wider variety of harmful behaviors.


By comparison, there is no relationship between personality level and harm for three of the Big-Five traits: users write 90\% or more safe messages regardless of their level of openness, extraversion, and neuroticism. The remaining two traits are conscientiousness and agreeableness, where users with lower-outlier trait levels write a higher ratio of unsafe messages. While we did not expect this result, in retrospect it seems unsurprising: users with low agreeableness levels do not get along well with others and thus write more unsafe messages, while users with low conscientiousness levels are more spontaneous and less filtered in their responses, and therefore seem to write in a way society finds less acceptable.

\begin{table}[htbp]
\scalebox{0.90}{
\centering
\begin{tabular}{cccc}
\hline
 Personality & Box distribution & 0 to 3 & 5 to 7 \\
\hline
 & Upper outliers & 83.5\% & 16.5\% \\ 
Na & Boxplot & 91.6\% & 8.4\% \\ 
 & Lower outliers & 98.0\% & 2.0\% \\ 
\hline
 & Upper outliers & 79.6\% & 20.4\% \\ 
Ma & Boxplot & 91.6\% & 8.4\% \\ 
 & Lower outliers & 95.6\% & 4.4\% \\ 
\hline
 & Upper outliers & 82.5\% & 17.5\% \\ 
 Ps & Boxplot & 91.6\% & 8.4\% \\
 & Lower outliers & 95.7\% & 4.3\% \\
\hline
 & Upper limit & 90.6\% & 9.4\% \\ 
Op & Boxplot & 91.6\% & 8.4\% \\ 
 & Lower limit & 95.8 \% & 4.2\% \\ 
\hline
 & Upper outliers & 93.5\% & 6.5\% \\ 
Co & Boxplot & 91.6\% & 8.4\% \\ 
 & Lower outliers & 69.7\% & 30.3\% \\ 
\hline
 & Upper outliers & 90.6\% & 9.4\% \\ 
Ex & Boxplot & 91.6\% & 8.4\% \\ 
 & Lower outliers & 95.7\% & 4.3\% \\ 
\hline
 & Upper outliers & 94.5\% & 5.5\% \\ 
Ag & Boxplot & 91.6\% & 8.4\% \\ 
 & Lower outliers & 84.5\% & 15.5\% \\ 
\hline
 & Upper outliers & 95.6\% & 4.4\% \\ 
Ne & Boxplot & 91.6\% & 8.4\% \\ 
 & Lower outliers & 89.5\% & 10.5\% \\ 
\hline
\end{tabular}
}
\caption{Percentage of safe and harmful messages per personality trait level in the Twitter dataset.}
\label{tbl:Perc-toxicMess-Twitter}
\end{table}

It is worth noting that Community Sift, like our study, is based on identifying and weighting keywords from the messages. This begs the question of whether both systems are picking out the same keywords in their analyses and giving correlated results. To begin, we can highlight that both systems handle the English language differently: LIWC sorts words semantically while Community Sift classifies them by harmfulness level, and LIWC includes non-word tokens such as punctuation marks which are not part of Community Sift. Focusing on vocabulary, we observe that, out of 191,097 different tokens in our random Twitter corpus, 53,757 are words recognized by the Community Sift system and 17,460 are recognized by LIWC, but the overlap between these sets is only of 9,783 words. This limited overlap indicates that most of the ranking done by each system is computed based on words ignored by the other. Furthermore, we can consider whether the 9,783 words recognized by both systems are rated similar risk levels, by checking whether words rated in the 0-3 and 5-7 levels were correlated positively or negatively, respectively, with Dark Triad traits. Overall, we find that the two systems disagree on the risk level of 34\% of the words they both recognize, meaning that either Community Sift rates them as safe but they correlate positively to a Dark Triad trait, or that Community Sift rates them as high-risk but they correlate negatively to a Dark Triad trait. Finally, we ran two more ``cleaned-up'' experiments. In the first, we filtered out from our Twitter dataset all words that Community Sift tags as levels 2-3 (mostly safe) and 5-7 (risky), to keep only levels 0-1 (very safe) and 4 (unknown), and ran this cleaned-up dataset through our system. In the second experiment, we determined which LIWC categories contain the greatest proportion of 5-7 words - they are the ``anger'', ``biological process'', ``body'', ``sexual'' and ``swear words'' categories, with between 28\% and 83\% risky words - and eliminate these categories entirely from our system. In both experiments, we still find the same relationship between Community Sift risk levels and predicted personality trait distributions as in Table \ref{tbl:Perc-toxicMess-Twitter}, although the precise percentages naturally vary a bit. This confirms that our system and Community Sift operate in mostly independent manners and with limited overlap, and they are not simply classifying users based on the same detected keywords.

\subsection{Number of Tweets} 
We performed an experiment to verify the minimum number of messages needed to make a correct personality prediction with our model. Indeed, our original target value of 400 tweets per user was selected as an arbitrary high number. In fact, for any practical moderation purpose, it would be necessary to predict a user's personality before they have posted such a high number of messages, as a harmful user will likely have caused harm to the community long before they hit the 400-message mark! Consequently, we want to determine how accurate a prediction we can obtain with fewer messages. For this experiment, we will assume that the individual's personality as computed at 400 messages is correct, and compute the correlation between that value and the personality computed with 1, 5, 10, 25, 50, 75, 100, 200, and 300 messages. We always use the messages in chronological order; 1 message thus means the oldest message in that user's history, while 300 messages exclude the 100 most recent messages the user posted. It is important to note that this will tell us how \textit{correlated} the values are, not how \textit{exact} they are. It means that, for example, a user with a high value for a personality trait when computed with 400 messages will likely also have a high value for that personality trait when computed with a few messages, but the exact personality trait value may vary. 

The correlation results are presented in Table \ref{tbl:messagecount}. It can be seen that the correlation between 1 message and 400 is rather low, but it increases quickly initially as the number of messages considered increases, to plateau at around 75 to 100 messages. With a set of 100 messages, all eight personality traits will have values with a correlation between 0.8 and 0.9 to their counterparts computed with 400 messages. Increasing the number of messages considered beyond 100 gives little improvement. Moreover, the results in Table \ref{tbl:messagecount} show that this observation holds for all eight personality traits. In fact, the correlation values increase in lockstep over all personality traits; there isn't a single trait that is easier or more difficult than the others to model. It thus appears that 100 messages are sufficient to gain a very good idea of a user's personality. At an average of 13.5 words and 85 characters per tweet, 100 messages represent 1,350 words or 8,500 characters.

For comparison, another study \cite{arnoux201725} has explored the number of tweets needed to predict a user's personality. Their results showed that a correlation begins to appear at 25 tweets, and stabilizes at 100 tweets. This is consistent with our results.

\begin{table}[htbp]
\scalebox{1.00}{
\begin{tabular}{lccccccccc}
\hline
Trait & 1 & 5 & 10 & 25 & 50 & 75 & 100 & 200 & 300 \\
\hline
Na & 0.14 & 0.30 & 0.43 & 0.60 & 0.70 & 0.78 & 0.84 & 0.92 & 0.98 \\
Ma & 0.23 & 0.41 & 0.54 & 0.72 & 0.82 & 0.87 & 0.90 & 0.96 & 0.99 \\
Ps & 0.21 & 0.38 & 0.49 & 0.70 & 0.80 & 0.86 & 0.89 & 0.96 & 0.99 \\
Op & 0.12 & 0.22 & 0.40 & 0.62 & 0.72 & 0.78 & 0.83 & 0.92 & 0.98 \\
Co & 0.21 & 0.44 & 0.57 & 0.75 & 0.83 & 0.88 & 0.91 & 0.97 & 0.99 \\
Ex & 0.14 & 0.25 & 0.43 & 0.61 & 0.70 & 0.77 & 0.81 & 0.92 & 0.98 \\
Ag & 0.19 & 0.31 & 0.45 & 0.65 & 0.76 & 0.84 & 0.88 & 0.94 & 0.98 \\
Ne & 0.16 & 0.47 & 0.61 & 0.75 & 0.82 & 0.87 & 0.90 & 0.96 & 0.99 \\
\hline
\end{tabular}
}
\caption{Correlation of personality score to 400 messages.}
\label{tbl:messagecount} 
\end{table}


\subsection{Reddit Experiment}\label{redditX}
\begin{figure}

\centering
\includegraphics[scale=0.5]{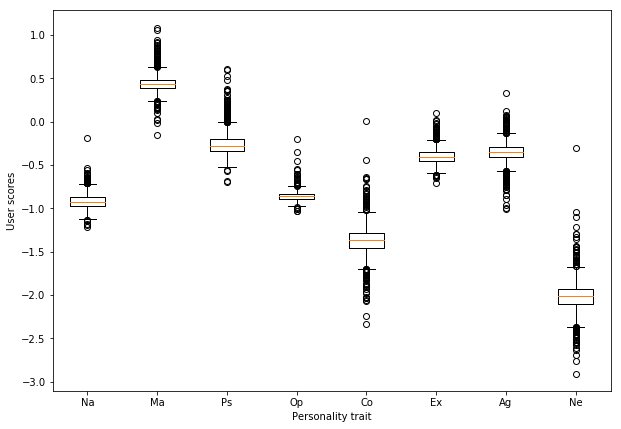}
\caption{\label{fig:box_reddit} Box plot representation of personality traits in Reddit dataset.}
\end{figure}

While the original correlations of \cite{sumner2012predicting} were computed on Twitter data, there is nothing in our methodology that is inherently limited to that social network. We thus expect that our methodology should work when applied in other online communities. To verify that assumption, we conducted a second experiment, using anonymized data from Reddit\footnote{https:\/\/www.reddit.com\/r\/bigquery\/comments\/3cej2b/ 17\_billion\_reddit\_comments\_loaded\_on\_bigquery}. Since Reddit does not have a message length limit like Twitter, each message is considerably longer; on average of 37 words and 220.6 characters. This means that users generate content approximately equal to 100 tweets using approximately only 40 messages, and consequently that is what we used as our threshold minimum number of messages per user. This gives us a dataset of 1,634 users who contributed 40 messages or more. We compute the eight personality trait scores of each user with the methodology described in Section \ref{methodology}, and show the spread of results in the boxplot of Figure \ref{fig:box_reddit}. Compared to Figure \ref{fig:box_twitter8}, it can be observed that the distribution of personalities are very similar, perhaps an indication that the same profile of individuals contributes to both communities. There are visibly more outlier users in each personality type than there were on Twitter, but that may simply be the result of having twice as many users in the dataset.

\begin{table}[h!]
\scalebox{0.90}{
\centering
\begin{tabular}{cccc}
\hline
 Personality & Box distribution & 0 to 3 & 5 to 7 \\
\hline
 & Upper outliers & 87.6\% & 12.4\% \\ 
Na & Boxplot & 90.5\% & 9.5\% \\ 
 & Lower outliers& 93.7\% & 6.3\% \\ 
\hline
 & Upper outliers & 86.5\% & 13.5\% \\ 
Ma & Boxplot & 90.5\% & 9.5\% \\ 
 & Lower outliers & 93.5\% & 6.5\% \\ 
\hline
 & Upper outliers & 85.6\% & 14.4\% \\ 
Ps & Boxplot & 91.5\% & 8.5\% \\ 
 & Lower outliers& 95.9\% & 4.1\% \\ 
\hline
 & Upper outliers & 92.5\% & 7.5\% \\ 
Op & Boxplot & 90.5\% & 9.5\% \\ 
 & Lower outliers & 91.8\% & 8.2\% \\ 
\hline
 & Upper outliers & 93.6\% & 6.4\% \\ 
Co & Boxplot & 90.5\% & 9.5\% \\ 
 & Lower outliers & 86.5\% & 13.5\% \\
\hline
 & Upper outliers & 89.6\% & 10.4\% \\ 
Ex & Boxplot & 90.5\% & 9.5\% \\ 
 & Lower outliers& 92.6\% & 7.4\% \\ 
\hline
 & Upper outliers & 93.6\% & 6.4\% \\ 
Ag & Boxplot & 90.5\% & 9.5\% \\ 
 & Lower outliers& 86.5\% & 13.5\% \\ 
\hline
 & Upper outliers & 92.5\% & 7.5\% \\ 
Ne & Boxplot & 90.5\% & 9.5\% \\ 
 & Lower outliers& 87.6\% & 12.4\% \\ 
\hline
\end{tabular}
}
\caption{Percentage of safe and harmful messages per personality trait in the Reddit dataset.}
\label{tbl:Perc-toxicMess-Reddit} 
\end{table}

We perform the same study of harmful messages per personality trait as for the Twitter dataset. The results, which are presented in Table \ref{tbl:Perc-toxicMess-Reddit}, are very similar to those obtained with the Twitter users. In particular, we still find that users with upper-outlier levels of Dark Triad traits or lower-outlier levels of agreeableness or conscientiousness traits post a greater proportion of unsafe messages and a lesser proportion of safe messages, and that openness and extraversion are not indicators of harmfulness. The only notable difference is that users with a lower-outlier level of neuroticism post a higher level of harmful messages, which was not the case in the Twitter dataset. Nonetheless, this experiment shows that, although Reddit and Twitter are different communities with differing themes and restrictions, the personalities can be modeled in the same way and correlate to harmfulness in the same manner.


\subsection{Subreddit Experiment}\label{subreddits}
As an additional experiment on Reddit, we decided to apply our personality prediction method on two well-known Reddit communities in terms of harm level, the r/DIY\footnote{https://www.reddit.com/r/DIY/} and r/The\_Donald subreddits. The r/DIY subreddit is one of the most important DIY (Do-It-Yourself) communities on Reddit. Created in 2008, it is currently active and counts more than 17.8M subscribers. It is a well-known positive space for showing ideas and sharing tips on home-made projects. By contrast, r/The\_Donald is a subreddit created in 2015 to support the 2016 presidential campaign of Donald Trump. At its peak, this subreddit was one of Reddit's most active communities and counted more than 800,000 subscribers. The subreddit was notorious for its hateful messages, and was ultimately shut down in June 2020 for not complying with Reddit regulations due to huge amounts of racism, violence, and harassment\footnote{https://www.nytimes.com/2020/06/30/us/politics/reddit-bans-steve-huffman.html}.

\begin{figure}
\centering
\includegraphics[scale=1.5]{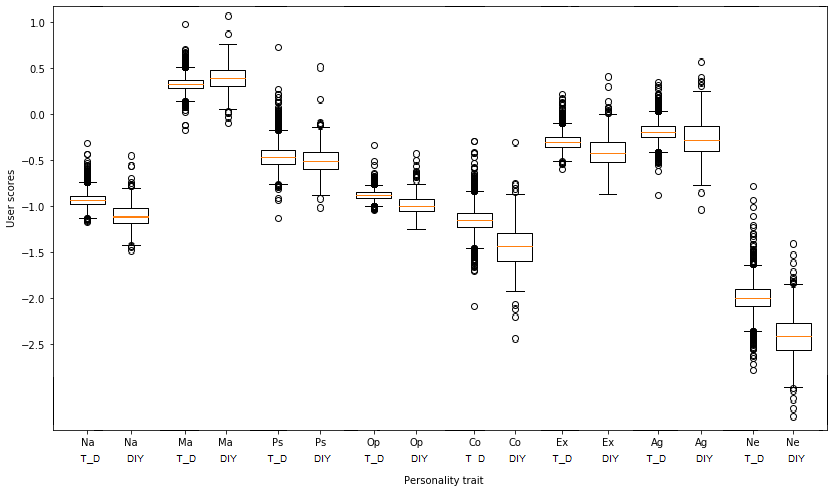}
\caption{\label{fig:box_reddit_combined} Boxplot representation of personality traits in the subreddits r/The\_Donald (T\_D) and r/DIY (DIY).}
\end{figure}

We obtain archives of these two subreddits from January to October 2018, and extract from them all users who posted more than 40 messages in that period. This yields 4,750 users from r/The\_Donald and 450 users from r/DIY. From our earlier observations of the data, there are actually many more active users in r/DIY, but each one posts only a few messages, while the fewer users of r/The\_Donald write a lot more messages each. In order to compare the two communities together, we perform the same study of message harmfulness per personality trait as before on each dataset separately. The personality score boxplots are presented 
side by side in Figure \ref{fig:box_reddit_combined}, 
while the safe and harmful message ratios per personality value are presented in Tables \ref{tbl:Perc-toxicMess-theDonald} and \ref{tbl:Perc-toxicMess-diy}. 

Looking first at the ratio of harmful messages per personality trait value in Tables \ref{tbl:Perc-toxicMess-theDonald} and \ref{tbl:Perc-toxicMess-diy}, it can be seen that users in r/The\_Donald post a greater proportion of harmful messages than the users in the r/DIY group and the random user sample used in Section \ref{redditX} across almost all personality traits and values. By contrast, users in the r/DIY subreddit post a lower proportion of harmful messages across all personality traits compared to both other groups, and sometimes post less than half as many harmful messages as r/The\_Donald users with the same personality trait. 
In terms of specific personality traits, the same pattern observed in the baseline Reddit community is still present in both specific communities: users with upper-outlier values of Dark Triad traits and lower-outlier values of consciousness, agreeableness, and neuroticism show a greater tendency to post harmful messages. However, the difference between the rate at which boxplot-users and outlier-users post harmful messages is much higher in the r/The\_Donald community than in the baseline community, while that rate is mostly equivalent between the r/DIY community and the baseline.
These observations confirm the prevalence of harmful messages in the r/The\_Donald subreddit and the healthy nature of the r/DIY community. More generally, they also demonstrate how our methodology can be used to compare specific sub-communities to a baseline community sample in order to measure their relative health. 

Comparing next the boxplot personality results, it can be seen that the general shape of both distributions of Figure \ref{fig:box_reddit_combined} remains consistent with that of the general Reddit community shown in Figure \ref{fig:box_reddit}. One difference we can observe between the communities is that the r/DIY community shows larger IQR boxes and extrema bounds as well as fewer distant outliers than either the baseline Reddit community or the r/The\_Donald community for all personality traits. This result indicates that the r/DIY community consists of a much more diverse set of users and personalities than average. 
The r/DIY community also appears to have lower scores in narcissism and neuroticism than either the baseline or the r/The\_Donald community, possibly reflecting the type of individual attracted to a DIY hobby group, and both r/DIY and r/The\_Donald have slightly higher agreeableness levels than the baseline, possibly reflecting cooperation in these like-minded special-interest communities. 

Taken together, the results of this experiment demonstrate that our personality prediction methodology can distinguish between different communities with different cultures and attitudes towards harmfulness. In future works, aggregated personality scores could be used as an indicator of the health of a community.



\begin{table}[h!]
\scalebox{0.90}{
\centering
\begin{tabular}{cccc}
\hline
 Personality & Box distribution & 0 to 3 & 5 to 7 \\
\hline
 & Upper outliers & 85.7\% & 14.3\% \\ 
Na & Boxplot & 89.7\% & 10.3\% \\ 
 & Lower outliers& 94.8\% & 5.2\% \\ 
\hline
 & Upper outliers & 82.7\% & 17.3\% \\ 
Ma & Boxplot & 89.7\% & 10.3\% \\ 
 & Lower outliers & 90.7\% & 9.3\% \\ 
\hline
 & Upper outliers & 82.5\% & 17.5\% \\ 
Ps & Boxplot & 90.6\% & 9.4\% \\ 
 & Lower outliers& 89.7\% & 10.3\% \\ 
\hline
 & Upper outliers & 88.5\% & 11.5\% \\ 
Op & Boxplot & 89.7\% & 10.3\% \\ 
 & Lower outliers & 94.8\% & 5.2\% \\ 
\hline
 & Upper outliers & 89.7\% & 10.3\% \\ 
Co & Boxplot & 89.7\% & 10.3\% \\ 
 & Lower outliers & 81.6\% & 18.4\% \\
\hline
 & Upper outliers & 89.8\% & 10.2\% \\ 
Ex & Boxplot & 89.7\% & 10.3\% \\ 
 & Lower outliers& 91.7\% & 8.3\% \\ 
\hline
 & Upper outliers & 93.8\% & 6.2\% \\ 
Ag & Boxplot & 89.7\% & 10.3\% \\ 
 & Lower outliers& 82.7\% & 17.3\% \\ 
\hline
 & Upper outliers & 89.7\% & 10.3\% \\ 
Ne & Boxplot & 89.7\% & 10.3\% \\ 
 & Lower outliers& 85.9\% & 14.1\% \\ 
\hline
\end{tabular}
}
\caption{Percentage of safe and harmful messages per personality trait in the subreddit r/The\_Donald.}
\label{tbl:Perc-toxicMess-theDonald} 
\end{table}

\begin{table}[h!]
\scalebox{0.90}{
\centering
\begin{tabular}{cccc}
\hline
 Personality & Box distribution & 0 to 3 & 5 to 7 \\
\hline
 & Upper outliers & 93.8\% & 6.2\% \\ 
Na & Boxplot & 95.8\% & 4.2\% \\ 
 & Lower outliers& 96.8\% & 3.2\% \\ 
\hline
 & Upper outliers & 90.5\% & 9.5\% \\ 
Ma & Boxplot & 95.8\% & 4.2\% \\ 
 & Lower outliers & 98.0\% & 2.0\% \\ 
\hline
 & Upper outliers & 93.8\% & 6.2\% \\ 
Ps & Boxplot & 95.8\% & 4.2\% \\ 
 & Lower outliers& 95.9\% & 4.1\% \\ 
\hline
 & Upper outliers & 93.9\% & 6.1\% \\ 
Op & Boxplot & 95.8\% & 4.2\% \\ 
 & Lower outliers & None & None \\ 
\hline
 & Upper outliers & 93.9\% & 6.1\% \\ 
Co & Boxplot & 95.8\% & 4.2\% \\ 
 & Lower outliers & 91.8\% & 8.2\% \\
\hline
 & Upper outliers & 99.0\% & 1.0\% \\ 
Ex & Boxplot & 95.8\% & 4.2\% \\ 
 & Lower outliers& None & None \\ 
\hline
 & Upper outliers & 98.0\% & 2.0\% \\ 
Ag & Boxplot & 95.8\% & 4.2\% \\ 
 & Lower outliers& 90.5\% & 9.5\% \\ 
\hline
 & Upper outliers & 93.9\% & 6.1\% \\ 
Ne & Boxplot & 95.8\% & 4.2\% \\ 
 & Lower outliers& 92.7\% & 7.3\% \\ 
\hline
\end{tabular}
}
\caption{Percentage of safe and harmful messages per personality trait in the subreddit r/DIY.}
\label{tbl:Perc-toxicMess-diy} 
\end{table}

\section{User Harm Prediction}\label{toxicityregression}
The previous sections showed that the features of a user's written text can be used to model their personality traits, and that personality traits are predictors of writing harmful messages. In this section, we explore the possibility of using regression algorithms to model and predict the harm level of users using only the 46 LIWC features of their written text we measured previously, without the intermediary step of modeling their personality traits. First, we need to define a baseline quantitative measure of the harm level of a user. To do this, we use the ratio of the number of high-risk messages (messages with risk levels 5 ($M_{u,5}$), 6 ($M_{u,6}$) and 7 ($M_{u,7}$) according to the Community Sift system) to the total number of user messages to be analyzed ($M_u$), as shown in Equation \ref{eq3}. 
Note that, while this equation uses all three risk levels equally, a more refined future version could weight them differently to compute a greater harm level for users that write higher-risk-level messages.

\begin{equation}\label{eq3}
T_u = \frac{M_{u,5} + M_{u,6} + M_{u,7}}{M_u}
\end{equation}

Using this formula, the harm level will always be between 0 and 1. However, to use a regression algorithm, the values must be continuous in space. The standard solution is to apply the $\Logit$ function which converts the values from $[0, 1]$ to $[-\infty, +\infty]$. The $\Logit$ function is defined in Equation \ref{eq4}. 
The complementary function to $\Logit$, which must be applied to reverse the process and analyze the results, is the $\Expit$ function, which is define in Equation \ref{eq5}

\begin{equation}\label{eq4}
\Logit(x) = \log(\frac{x}{1-x})
\end{equation}

\begin{equation}\label{eq5}
\Expit(x) = \frac{1}{1+\exp(-x)}
\end{equation}

\begin{figure}[htbp]
\centering
\includegraphics[scale=1.5]{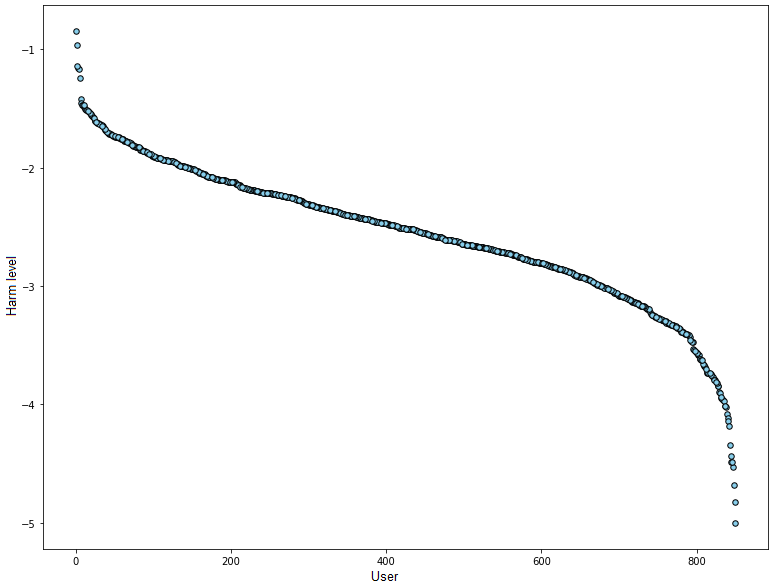}
\caption{Distribution of Twitter users sorted according to their harm level.}
\label{figTwitterToxLvl} 
\end{figure}

For reference, the harm level of the users in our random Twitter dataset, after applying the $\Logit$ function, is given in Figure \ref{figTwitterToxLvl}. 

We decided to use a variety of regression algorithms, in order to compare their results and determine which algorithm best suits our problematic and our datasets. The algorithms used are: Ridge regression, Ridge regression using cross-validation (RidgeCV), Bayesian Ridge Regression, Stochastic Gradient Descent Regression (SGD) and Huber Regression. All these regression algorithms are implemented in the Scikit-learn library \cite{scikit-learn,scikit-learn2}.

\subsection{Regrssion Results}\label{regressionresults}
For this experiment, we train each regression algorithm to discover a degree-3 polynomial using the 850 Twitter users of Section \ref{experimentTwitter} and we test them using the 1,634 Reddit users of Section \ref{redditX}. The level 3 polynomial was chosen because it corresponds visually to the shape of the distribution of Figure \ref{figTwitterToxLvl}. To measure the performance of each of the algorithms we use two metrics, the Pearson correlation and the mean squared error (MSE). The MSE gives a per-user error: it indicates how off from the baseline harm level of equation \ref{eq3} the level predicted by the regression is. By contrast, the correlation gives a distribution-wide error: it indicates how well the ordering of users by harm level obtained by the regression corresponds to the baseline ordering. The Pearson correlation and MSE results obtained by the five trained regression algorithms are presented in Table \ref{tbl4.1:corrPearson}. Looking at these results, it can be seen that all five regression algorithms give similar results. Therefore, any of these algorithms can be used for this purpose. The correlation is almost 0.5, meaning that the regression maintains roughly the ordering of users by harm level, and the MSE is less than 0.003, or 1\% of the harm level value. Taken together, these results indicate that the regression will have more difficulty ranking users by harm level in the centre of the distribution of Figure \ref{figTwitterToxLvl}, where users have very similar scores to each other and a small error can change a user's rank dramatically, but will be more confident of its rankings at the edges of the distribution, where high and low outlier values are found and a small error has little to no impact on ranking.

\begin{table}[!ht]
\centering
\begin{tabular}{l c c}
\hline
 & \textbf{Pearson correlation} & \textbf{MSE}
\\\hline
Ridge & 0.5238 & 0.0025
\\
RidgeCV & 0.5208 & 0.0024
\\
Bayes & 0.5241  & 0.0023
\\
SGD & 0.4486 & 0.0024 
\\
Huber & 0.4368 & 0.0028
\\\hline
\end{tabular}
\caption{Results of the Pearson correlation and MSE.}
\label{tbl4.1:corrPearson}
\end{table}

\subsection{Contribution of each linguistic feature}
\begin{figure}[htbp]
\centering
\includegraphics[scale=0.50]{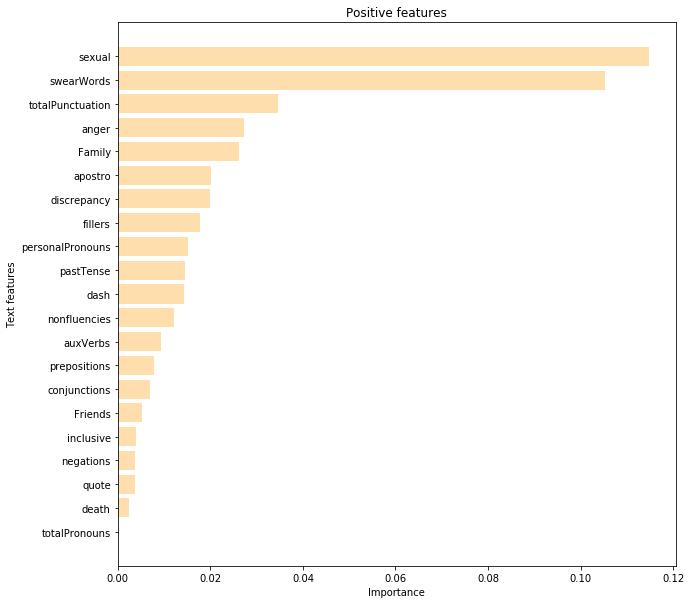}
\caption{Linguistic features that are positive predictors of harm level.}
\label{posFeat} 
\end{figure}

\begin{figure}[htbp]
\centering
\includegraphics[scale=0.50]{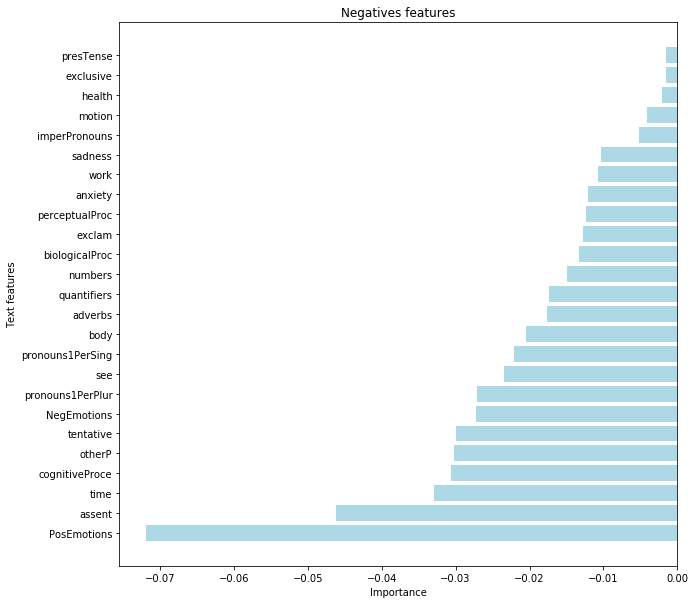}
\caption{Linguistic features that are negative predictors of harm level.}
\label{negFeat} 
\end{figure}

Next, we can determine which of the 46 linguistic features are the strongest positive and negative indicators of harm. If we had interpolated a degree-1 polynomial, we could do this simply by checking the value of the coefficient of each term of the polynomial, corresponding to each individual linguistic feature. In a degree-3 polynomial, however, the task is no so simple, as features will appear combined with others in multiple terms. Instead, we measure the contribution of each feature empirically using the following method. First, we select the set of Reddit users who have non-zero values for all features. This gives us a set of 702 users. For each one, we set the value of one linguistic feature to zero, recompute their harm level, measure the difference with the original harm level and divide it by the feature's value. This gives us the contribution per unit of that feature in the harm level of that user. We repeat this process for each of the 702 users to compute the average contribution of each feature, and for each of the 46 features in our study. We present in Figures \ref{posFeat} and \ref{negFeat} the features that are found to be positive and negative contributors to the harm level, respectively.

In can be seen from these figures that certain linguistic features have much more influence on a user's harm level than others. Sexual words (sex, ass, horny, etc.) and swear words (fuck, damn, shit, etc.) are by far the two features that are the greatest positive predictors harm level, while positive emotions (love, joy, happy, etc.) and, to a lesser extent, assent words (agree, accept, yeah, etc.) are the greatest negative predictors. These results are coherent with those found by other authors. In \cite{sumner2012predicting}, sexual words have the third highest positive correlation with narcissism and swear words have the strongest positive correlation to Machiavellianism and psychopathy, while positive emotions have the strongest negative correlation with Machiavellianism and psychopathy. Likewise, the authors of \cite{preotiuc2016studying} found that the linguistic attributes most strongly correlated with Dark Triad traits were swear words and porn words, the latter being a linguistic feature created by the authors of that study but closely related to the LIWC sexual words category. 


\subsection{Relationship between Harm Level and Personality}
\begin{figure}[htbp]
\centering
\includegraphics[scale=1.5]{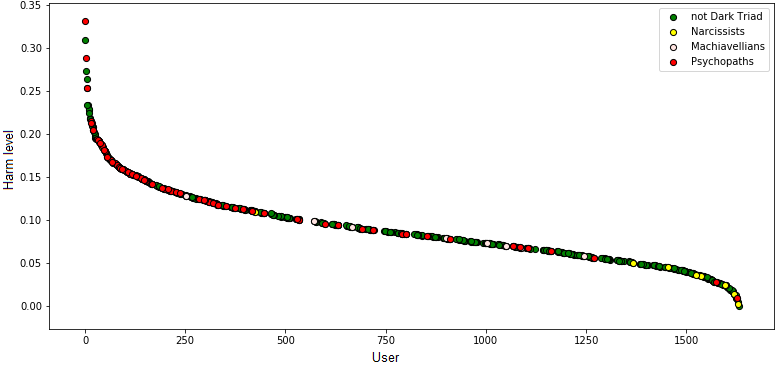}
\caption{Reddit users sorted according to their harm level computed using formula \ref{eq3}, with Dark Triad traits colored.}
\label{BRToxicDist} 
\end{figure}

\begin{figure}[htbp]
\centering
\includegraphics[scale=1.5]{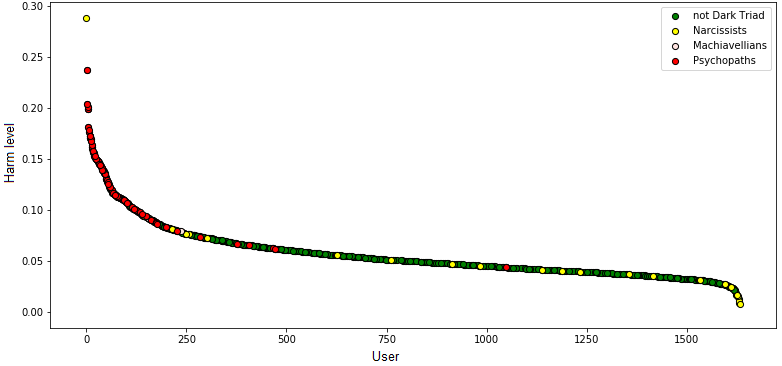}
\caption{Reddit users sorted according to their harm level obtained with the Bayesian Ridge regression algorithm, with Dark Triad traits colored.}
\label{TLToxicDist} 
\end{figure}

\begin{figure}[!ht]
\centering
\includegraphics[scale=1.5]{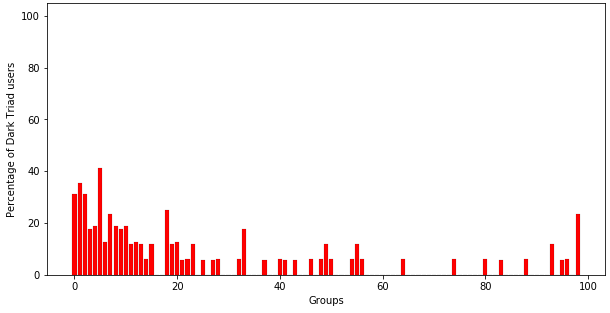}
\caption{Percent of Dark Triad users per percentile group of Figure \ref{BRToxicDist}.}
\label{centile4.4}
\end{figure}

\begin{figure}[!ht]
\centering
\includegraphics[scale=1.5]{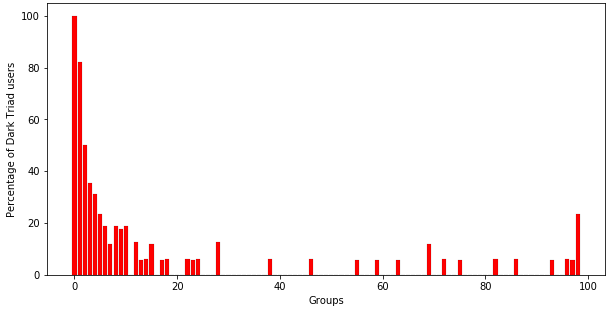}
\caption{Percent of Dark Triad users per percentile group of Figure \ref{TLToxicDist}. }
\label{centile4.5}
\end{figure}

Finally, we can compare the harm levels, both the baseline ones and the regression predictions, with the personality prediction results of Section \ref{redditX}. We ordered users by their harm level calculated with Equation \ref{eq3} in Figure \ref{BRToxicDist} and with the Bayesian Ridge regression algorithm in Figure \ref{TLToxicDist}. In both graphics, we mark users with upper-outlier narcissism scores in yellow, Machiavellianism in pink, psychopathy in red (including those with high values of both psychopathy and Machiavellianism), and users that are not upper-outliers in any Dark Triad traits in green. To further understand these figures, we also divided the distributions into percentiles, keeping the users in the same harm-level order as in Figures \ref{BRToxicDist} and \ref{TLToxicDist}, and counted the percentage of users with at least one upper-outlier Dark Triad trait in each percentile group. These statistics are shown in Figures \ref{centile4.4} and \ref{centile4.5}.

It can be seen from these results that Dark Triad users, especially those with high scores in psychopathy and Machiavellianism, are also those found to have high harm levels using Equation \ref{eq3} and predicted to have high harm levels by our regression algorithm. However, high psychopathy and Machiavellianism users are more spread out across the entire distribution in Figures \ref{BRToxicDist} and \ref{centile4.4}, and are more concentrated in the high-harm-level portion of the distribution in Figures \ref{TLToxicDist} and \ref{centile4.5}. This indicates that our regression algorithm's prediction of which users will have a high harm level is coherent with our personality model's prediction of which users have Dark Triad personalities. 

It is also interesting to note that, unlike psychopathy and Machiavellianism, high narcissism users are spread out across the distribution in both figures. This reflects the fact that a high level of narcissism is a much weaker indicator of harmful behavior than the other two Dark Triad traits, as we found in Tables \ref{tbl:Perc-toxicMess-Twitter}, \ref{tbl:Perc-toxicMess-Reddit}, \ref{tbl:Perc-toxicMess-theDonald} and \ref{tbl:Perc-toxicMess-diy}, and as indicated by other authors \cite{buckels2014trolls}, \cite{preotiuc2016studying} we discussed previously. This result confirms that our regression method is indeed predicting harm levels, and not just Dark Triad traits.

However, it can also be seen that some of the users with the highest harm levels in Figure \ref{BRToxicDist} are users without any high Dark Triad traits. These users use ordinary language and their values of the measured features are not notable, therefore they mislead both our personality prediction method and the harm level prediction, but get caught for having written harmful messages by Community Sift. This illustrates a limitation of both our methodologies of Section \ref{methodology} and of Section \ref{toxicityregression}: since they rely on linguistic features exclusively, they can be fooled into giving a low Dark Triad rating or harmfulness level to a user who writes, either accidentally or deliberately, in an innocuous way.

\section{Ethical Considerations}\label{danger} 
In this research, we model users' personality traits and to predict which ones are more likely to post harmful comments online. Clearly, such work raises several ethical questions that should be considered.

First of all, it is important to emphasize that the objective of our work is not to create a tool for psychological diagnostic. While we use psychological labels in the form of the Big-Five and Dark Triad personality traits, we only assign them to users based on statistical correlations and not through any formal psychological testing. In essence, our method only measures how similar a user's observed behavior is to that of other users with a specific personality trait. It does not accurately measure the user's actual personality and its results should not be mistaken for those of a legitimate psychological test.

In addition, our method can only predicts that a user is more or less likely to post harmful comments, it does not guarantee that they will or will not post such comments. Moderators should not take preemptive actions based on that prediction, as they may end up penalizing innocent users and harm rather than help their community.

It is important to note as well that communities that wish to use a tool based on our work must do so in accordance with all applicable regulations. For example, the GDPR states that users have a right to know of automated online profiling software, of their internal logic and of the uses made of their results \cite{regulation2016regulation}. We strongly advise that communities be forthcoming with users about the use of such tools and obtain their consent.

Furthermore, an important source of variation in the characteristics of written text comes not from personality but from cultural and racial background. This means that a personality model and harm level regression trained using messages written by a culturally or racially homogeneous group of users will correctly pick out harmful individuals from that group, but will make wrong predictions about users from other groups, including by mislabeling them as harmful at a disproportionate rate \cite{sap2019risk}. Care must thus be taken to train these systems using as varied a pool of users as possible. In this respect, neither \cite{sumner2012predicting} nor any of the other studies presented in Section \ref{relwork} have given the demographic or cultural make-up of the population of users they studied, and to the best of our knowledge there exists no social network message dataset that includes demographic information of the respondents.

\section{Conclusion} \label{conclusion}
In this paper, we presented a new way to predict the Dark Triad and Big-Five personality traits of online users. Based on the study of \cite{sumner2012predicting}, we created a mathematical model to predict these eight personality traits based on their correlation to writing style and vocabulary use. We tested our model on two datasets, one from Twitter and the other one from Reddit, and contrasted these results with a baselines harm measure of the messages posted. Our results showed that users with high narcissism, Machiavellianism, and psychopathy scores and lower agreeableness and conscientiousness scores write more unsafe messages. These results are consistent across both datasets, showing our method can be applied on different online communities. We also showed that aggregating results for all active users in a community can give a perspective on the overall health and harmfulness of that community. Next, we presented a technique to estimate the level of harm of users using the linguistic features of their written text and different regression algorithms. Our results show that it is indeed possible to predict the harm level of users by analyzing the different characteristics of the messages they post online. We have also found that most of the users that our models predict to have high harm levels are also those that our personality prediction model designates to have very high Dark Triad values.

The long-term goal of our project is to use this information to create tools that can help monitor and maintain the health of online community, by allowing community moderators to focus their resources on users that are predicted as more likely to post harmful comments. 

Although the work of \cite{sumner2012predicting} provides a thorough and detailed list of correlations between language attributes and personality, which is why it was chosen as a basis for our work, other similar studies have been done by other authors, such as \cite{preotiuc2016studying,schwartz2013personality}. These papers offer alternative correlation values that, while mostly consistent with those of \cite{sumner2012predicting}, do show many minor differences and some interesting important divergences. One direction for future work will be to analyze in depth the differences found in these studies, and to use them to create a more accurate list of correlations. This, in turn, should allow our model to more accurately model the personality traits of users. Similarly, the authors of \cite{arnoux201725} reported more success predicting personality traits using word embeddings rather than word counts. Consequently, we could enhance our model by using word embeddings to detect words with similar contextual meaning to those listed in LIWC, instead of being limited to its word lists.

An interesting additional direction for future work could be to have volunteers fill out self-reporting personality questionnaires, and compare their personality scores with those predicted by our method. This would give us an objective benchmark for our results and allow us to measure the overall accuracy of our metrics.

Finally, we have made a simplifying assumption in our work, namely to consider all harmful messages as part of a single ``harmful behavior'' group. In fact, different harmful behaviors manifest in very different ways. For instance, cyberbullying is a very different behavior from sexting or from grooming, and will be carried out using very different styles of messages. Thus we can assume that the written text characteristics of messages from different forms of harmful behaviors will be different, and that these difference will lead to variations in personality trait values and in the regression formula for users engaging in each behavior. While our work in this paper has demonstrated that it is possible to measure personality traits and general harm level from written text characteristics, future work will have to study the differences between various harmful behaviors in order to fine-tune our methodology and results. 

\section*{Acknowledgment}
This research was made possible by the financial, material, and technical support of Two Hat Security Research Corp., including in particular free unlimited access to the Community Sift harm labeling system, as well as the financial support of the Canadian research funding agency MITACS grant number IT09326.


\bibliographystyle{acl_natbib}
\bibliography{biblio}

\end{document}